\documentclass[lettersize,journal]{IEEEtran}

\usepackage{algorithm}
\usepackage{array}
\usepackage[caption=false,font=normalsize,labelfont=sf,textfont=sf]{subfig}
\usepackage{textcomp}
\usepackage{stfloats}
\usepackage{url}
\usepackage{verbatim}
\usepackage{graphicx}
\usepackage{cite}
\usepackage{xcolor}
\usepackage{mathtools}
\usepackage{epsfig,makeidx,color}
\usepackage{subcaption}
\usepackage{caption}
\usepackage{placeins}
\usepackage{amsmath,amssymb,bbm,enumitem,amsthm,amsfonts}
\usepackage{algorithm}
\usepackage{algcompatible}
\usepackage{array}
\usepackage{cite,graphicx,lipsum}
\usepackage{makecell, array, tabularx, multicol, multirow, booktabs, boldline}

\usepackage[switch,pagewise]{lineno}
\usepackage{hyperref}
\usepackage[cjk]{kotex}
\hypersetup{
        colorlinks = true,
        citecolor=red,
}

\def\cX{{\cal X}}

\def\cV{{\cal V}}

\def\be{ \begin{equation} }
\def\ee{ \end{equation} }
\def\bea{ \begin{eqnarray} }
\def\eea{ \end{eqnarray} }
\def\bt{{\bf t}}
\def\bx{{\bf x}}
\def\by{{\bf y}}

\def\bb{{\bf b}}

\def\bs{{\bf s}}

\def\bz{{\bf z}}

\def\be{{\bf e}}

\def\b0{{\bf 0}}

\ifCLASSOPTIONonecolumn
\else

\fi

\usepackage{fancyhdr}

\fancypagestyle{preprintfirstpage}{
    \fancyhf{}
    \fancyhead[C]{%
        \scriptsize
        This work has been submitted to the IEEE for possible publication. Copyright may be transferred without notice, after which this version may no longer be accessible.
    }
    
}

\begin{document}

\title{Dual-Stage Channel-Posterior Integration for MLM-Aided Token Communications}

\author{Hyelin Nam, Jihoon Lee, Seung-Woo Ko, Jinho Choi, and Seong-Lyun Kim
\thanks{H. Nam, J. Lee, and S.-L. Kim are with the School of Electrical
and Electronic Engineering, Yonsei University, Seoul, South Korea
(e-mail: \{hlnam, jhlee\}@ramo.yonsei.ac.kr, slkim@yonsei.ac.kr).}
\thanks{S.-W. Ko is with the Department of Smart Mobility Engineering,
Inha University, Incheon, South Korea
(e-mail: swko@inha.ac.kr).}
\thanks{J. Choi is with the School of Electrical and Mechanical Engineering,
Adelaide University, Adelaide, Australia
(e-mail: jinho.choi@adelaide.edu.au).}
}

\maketitle
\thispagestyle{preprintfirstpage}
\begin{abstract}
Token transmission over conventional digital modulation is vulnerable to channel impairments, leading to bit errors that cause received tokens to become contextually inconsistent. This paper proposes a receiver-side dual-stage channel-posterior integration framework for masked language model (MLM)-aided token reconstruction. In the proposed framework, the receiver detects unreliable token positions from channel reliability and constructs a vocabulary-level channel-posterior.
In particular, within the receiver, Posterior-Weighted Input Embedding (PWE) injects this posterior into the MLM input, while Adaptive Output Logit Fusion (AOF) reuses it at the output with a channel-adaptive weight. Consequently, multiple unreliable tokens can be reconstructed in a single bidirectional MLM inference without retransmission, transmitter-side encoding, or additional MLM training. Experiments over noisy channels show up to a $67.2$\% improvement in cosine similarity over context-only MLM reconstruction, with consistent gains across different MLM architectures and existing language-model-aided methods.
\end{abstract}

\section{Introduction}

Recent advances in generative and language models (LMs) have established tokens as fundamental units, since tokens represent words or subwords as compact discrete representations with contextual meaning. This has motivated token communication, where text is transmitted as a sequence of tokens over wireless networks to enable context-aware transmission and recovery through a language interface.

Since tokens are not designed for transmission robustness, channel-induced errors can introduce contextually inconsistent tokens and distort sentence meaning \cite{lee2026semantics, nam2024language}, motivating the use of language models (LMs) for reliable token delivery. Existing LM-based token reconstruction methods leverage large LMs (LLMs) to recover corrupted tokens from their linguistic context \cite{hao2025short, wang2026context, ren2025separate}. On the other hand, the linguistic ambiguity of text makes it difficult to recover corrupted tokens using only inter-token context. Several recent studies attempt to address it by combining the received signal’s reliability, known as channel likelihood, with the linguistic plausibility of candidate tokens estimated by LLMs \cite{wang2025large}, or masked LMs (MLMs) \cite{wang2026cl, wang2026icl, shin2026context}. In \cite{wang2026icl} and \cite{shin2026context}, for instance, the tokens are initially reconstructed by an MLM, and iteratively refined by incorporating the channel likelihood as posterior information. Such post-refinement processing using channel reliability is effective when the initial MLM-based reconstruction is relatively accurate. If not, early inaccurate reconstructions propagate through subsequent reconstruction steps, increasing computational burden due to extensive iterations.

This paper proposes a receiver-side text reconstruction framework without iterations by using the channel posterior in two different ways during a single inference. The receiver first identifies unreliable token positions from demodulation-derived channel reliability and selectively masks them. Posterior-Weighted Input Embedding (PWE) injects the channel-posterior into the MLM input, allowing channel information to directly guide the internal contextual inference rather than being considered only after prediction. Adaptive Output Logit Fusion (AOF) then reuses the same posterior to refine the MLM output, adaptively controlling its contribution according to channel quality. By incorporating channel information before inference, all unreliable positions can be jointly reconstructed by the MLM in a single inference, while AOF further resolves the remaining ambiguity at the output without iterative candidate-wise refinement. PWE and AOF provide complementary benefits across channel conditions, allowing the proposed framework to maintain robust reconstruction over a wide range of signal-to-noise ratios (SNRs).    

\begin{figure}[t]
  \centering
  \includegraphics[width=0.98\linewidth]{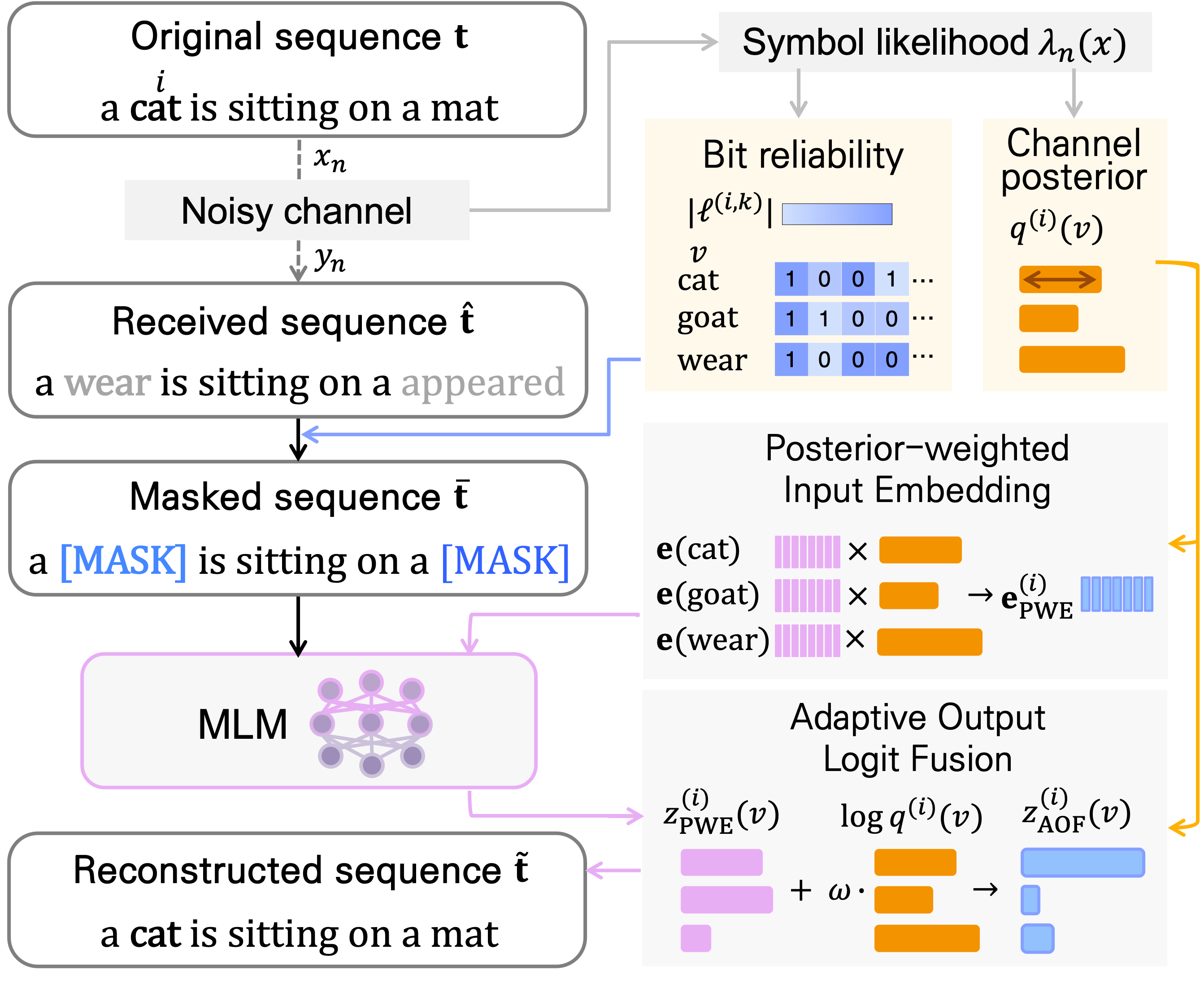}
  \caption{Overview of the Dual-Stage Channel-Posterior Integration for MLM-Aided text reconstruction framework.}
  \label{fig:main}
  \vspace{-0.5cm}
\end{figure}

The main contributions of this paper are summarized as follows:
\begin{itemize}
\item We develop a one-shot receiver-side framework that detects unreliable
tokens and jointly reconstructs them through a single bidirectional MLM
inference without iterative refinement.

\item We propose dual-stage channel-posterior integration, where PWE enables
channel-aware contextual inference across unreliable tokens and AOF adaptively
fuses the MLM output with the channel posterior according to channel quality.

\item We demonstrate complementary gains of PWE and AOF, enabling dual stage to achieve robust performance across SNRs, modulation schemes, and MLM architectures at lower complexity than iterative methods.
\end{itemize}
The remainder of this paper presents the system model in Section II, the proposed framework in Section III, numerical results in Section IV, and the conclusion in Section V.

\section{System Model}
Consider a point-to-point language transmission system, where a transmitter sends a natural language sentence to a single-antenna receiver through a wireless channel. A sentence is first converted into a sequence of token indices by a pre-trained tokenizer. Let $\mathcal{V}$ denote the vocabulary shared by the transmitter and the receiver, where $|\mathcal{V}|$ is the vocabulary size. The resulting token-index sequence is represented as,
\begin{equation} 
    \bt = [t^{(1)}, t^{(2)}, \ldots, t^{(L)}], \qquad t^{(i)} \in \mathcal{V}, 
\end{equation} 
where $L$ denotes the number of tokens for the sentence and $t^{(i)}$ represents the $i$th token index. Through a deterministic binary mapping function, denoted by $f_{\mathrm{b}}(\cdot)$, $t^{(i)}$ is then converted into a fixed-length bitstream of $B$-bits, given as $\bb^{(i)} = f_{\mathrm{b}}(t^{(i)}) \in \{0,1\}^{B}$, where $B \triangleq \left\lceil \log_2 |\mathcal{V}| \right\rceil$. By concatenating the bitstreams of all $L$ tokens, the sentence-level bit sequence is obtained as,
\begin{equation} 
    \bs = [\bb^{(1)},\bb^{(2)},\ldots,\bb^{(L)}] \in \{0,1\}^{L\cdot B},
\end{equation}
which is to be transmitted using a classical digital modulation scheme with modulation order $M$, where $M=2$ and $16$ correspond to BPSK and 16-QAM, respectively, by segmented into groups of $\log_2 M$ bits and mapped to complex-valued modulation symbols. This process is expressed as
\begin{equation} 
    \bx = f_{\mathrm{m}}(\bs) = [x_1,x_2,\ldots,x_N], \quad x_n\in\cX,
\end{equation} 
where $\cX$ is the signal constellation, $f_{\mathrm{m}}(\cdot)$ denotes the modulation mapping function and $N = \left\lceil \frac{L B}{\log_2 M} \right\rceil$ is the number of transmitted modulation symbols. If necessary, zero padding is applied to $\bs$ to make the total number of bits divisible by $\log_2 M$. The transmitter then transmits the complex-valued sequence through an additive white Gaussian noise (AWGN) channel. For the $n$-th transmitted symbol, the received signal is expressed as,
\begin{equation} 
    y_n = x_n + z_n, \qquad n=1,\ldots,N, 
\end{equation} 
where $z_n\sim\mathcal{CN}(0,N_0)$ is the additive white Gaussian noise, with noise power $N_0$.
Let $E_{\mathrm{s}}=\mathbb{E}[|x_n|^2]$ denote the average energy per modulation symbol, the bit-wise signal-to-noise ratio (SNR) is defined as
\begin{equation}
    \gamma_{\mathrm{b}}
    \triangleq
    \frac{E_{\mathrm{s}}}{N_0\log_2 M}.
\label{eq:bit_snr}
\end{equation}

At the receiver, the received symbol sequence $\by=[y_1,\ldots,y_N]$ is demodulated and demapped into a bit sequence using $f_{\mathrm{m}}^{-1}(\cdot)$, which maps each noisy symbol to the nearest valid constellation label. The resulting bit sequence is expressed as,
\begin{equation} \label{eq: demodul}
    \hat{\bs}=f_{\mathrm{m}}^{-1}(\by) =[\hat{\bb}^{(1)},\hat{\bb}^{(2)},\ldots,\hat{\bb}^{(L)}],
\end{equation}
where $\hat{\bb}^{(i)}\in\{0,1\}^{B}$ denotes the received bitstream corresponding to the $i$-th token. Due to the noisy channel, transmitted bits can be incorrectly recovered, resulting in bit flips between 0 and 1. Then the received token sequence is obtained by applying the inverse binary mapping to each received $i$-th bitstream, $\hat{t}^{(i)} = f_{\mathrm{b}}^{-1}(\hat{\bb}^{(i)})$, and is denoted by
\begin{equation} \label{eq: receivedseq}
    \hat{\bt}=[\hat{t}^{(1)},\hat{t}^{(2)},\ldots,\hat{t}^{(L)}], \qquad \hat{t}^{(i)}\in\cV.
\end{equation}
The bit errors make $\hat{\bt}$ linguistically inconsistent with $\bt$.
To address this, the following section presents a channel-aware MLM-aided
reconstruction framework that maps the received token sequence to the final
reconstructed sequence $\tilde{\bt}$.

\section{Dual-Stage Channel-Posterior Integration for MLM Reconstruction: PWE and AOF}

\subsection{Unreliable Token Detection for Masking Decision}
The receiver first identifies unreliable token positions to be reconstructed by the MLM using bidirectional context. Let $\bb^{(i)}=[b^{(i,1)},\ldots,b^{(i,B)}]$ denote the bitstream of the $i$-th token index. For each received symbol $y_n$, the receiver estimates the noise power $\hat{N}_0$ from $\mathrm{n}_{\mathrm{p}}$ number of pilot symbols, where $\hat{N}_0 \sim \mathrm{Gamma}(\mathrm{n}_{\mathrm{p}},N_0/\mathrm{n}_{\mathrm{p}})$~\cite{pauluzzi2000comparison, kay1993fundamentals}, and evaluates the symbol likelihood of each $x$ in $\cX$ of given modulation as
\begin{equation}
    \lambda_n(x)\triangleq p(y_n\mid x)= \exp\!\left(-\frac{|y_n-x|^2}{\hat{N_0}}\right) .
\label{eq:symbol_likelihood}
\end{equation}
The receiver then obtains the bit-level log-likelihood ratio (LLR) from the symbol likelihoods to evaluate whether each $k$-th bit of the $i$-th token, $b^{(i,k)}$, is more likely to be 1 or 0. Assume that $b^{(i,k)}$ is carried by the $n$-th modulation symbol, $x_n$. Let $\mathcal{X}^{(i,k)}(b)\subseteq\mathcal{X}$ denote the set of $n$-th transmitted symbols when $b^{(i,k)}=b$, where $b\in\{0,1\}$. Then, the LLR of $b^{(i,k)}$ is given by
\begin{equation}
    \ell^{(i,k)} \triangleq \log \frac{p(y_n\mid b^{(i,k)}=1)}{p(y_n\mid b^{(i,k)}=0)} =
    \log\frac{\sum\nolimits_{x\in\mathcal{X}^{(i,k)}(1)}\lambda_n(x)
    }{\sum\nolimits_{x\in\mathcal{X}^{(i,k)}(0)}\lambda_n(x)}.
\label{eq:bit_llr}
\end{equation}
The sign of $\ell^{(i,k)}$ determines the hard bit decision, while its magnitude $|\ell^{(i,k)}|$ represents the bit reliability. 
Since an error in any of the $B$ bits can change the decoded token index, we define the token-level reliability by the least reliable bit as
\begin{equation}
r^{(i)}=\min_{1\le k\le B}|\ell^{(i,k)}|, \quad i=1,\ldots,L.
\end{equation}
Using this score, the receiver defines a masking indicator based on whether
$r^{(i)}$ is below a pre-defined \textit{masking threshold} $\tau$:
\begin{equation}
    m^{(i)} =\begin{cases}
    1, & \text{if } r^{(i)} < \tau,\\
    0, & \text{if } r^{(i)} \ge \tau.
    \end{cases}
\end{equation}
Tokens with $m^{(i)}=1$ are regarded as \textit{unreliable} and reconstructed by the proposed MLM-aided process, while the remaining tokens are retained.

\begin{figure*}[!t]
\centering
\begin{minipage}[t]{0.47\textwidth}
\centering
\vspace{0pt}

\includegraphics[width=\linewidth]{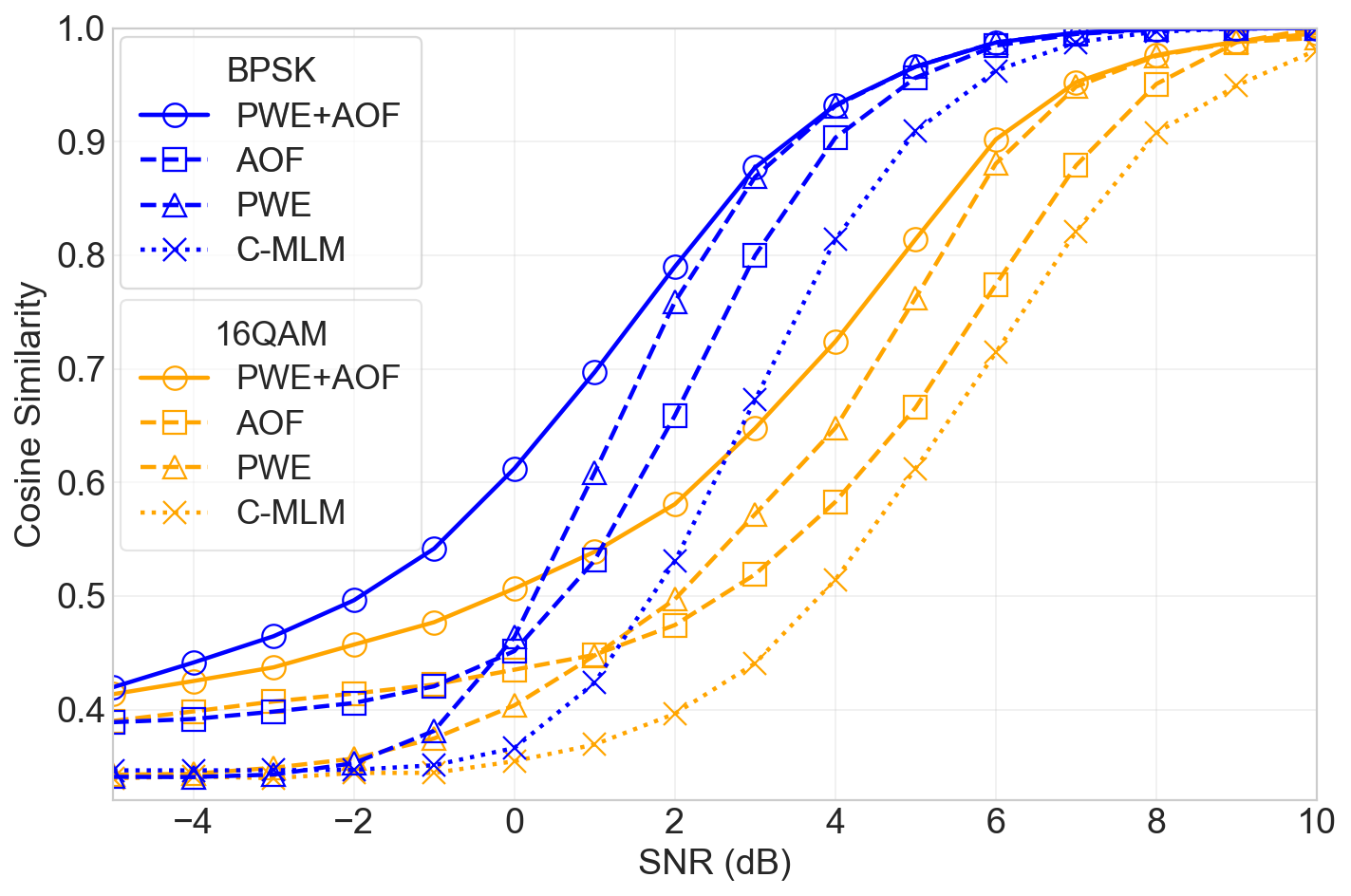}

\caption{Reconstruction performance using channel-aware reconstruction methods under different modulation schemes.}
\label{fig:modul}

\end{minipage}
\hfill
\begin{minipage}[t]{0.50\textwidth}
\centering
\vspace{0pt}

\captionof{table}{
Cosine Similarity across different MLMs using PWE, AOF and PWE+AOF under different SNR levels.
}
\label{tab:model}

\renewcommand{\arraystretch}{1.08}
\setlength{\tabcolsep}{8pt}
\small

\resizebox{\linewidth}{!}{%
\begin{tabular}{@{}l c c c c c@{}}
\toprule

\textbf{\makecell{MLM\\Method}}
& \makecell{SNR\\(dB)}
& \makecell{BERT-H128 \\ \cite{turc2019wellreadstudentslearnbetter}}
& \makecell{ALBERT \\ \cite{lan2019albert}}
& \makecell{ELECTRA \\ \cite{clark2020electra}}
& \makecell{RoBERTa \\ \cite{liu2019roberta}} \\

\midrule

\multirow{3}{*}{C-MLM}
& $-5$ & 0.4051 & 0.3667 & 0.4154 & 0.3378 \\
& $0$  & 0.4165 & 0.3727 & 0.4284 & 0.3586 \\
& $5$  & 0.8839 & 0.8135 & 0.9135 & 0.9256 \\

\cmidrule(lr){2-6}

\multirow{3}{*}{PWE}
& $-5$ & 0.4123 & 0.3667 & 0.4419 & 0.3602 \\
& $0$  & 0.5178 & 0.3886 & 0.5643 & 0.5354 \\
& $5$  & 0.9492 & 0.9454 & 0.9413 & 0.9641 \\

\cmidrule(lr){2-6}

\multirow{3}{*}{AOF}
& $-5$ & 0.4808 & 0.4351 & 0.4738 & 0.4485 \\
& $0$  & 0.5769 & 0.4636 & 0.6041 & 0.5246 \\
& $5$  & 0.9589 & 0.8613 & 0.9421 & 0.9700 \\

\cmidrule(lr){2-6}

\multirow{3}{*}{\makecell{PWE+AOF\\(Proposed)}}
& $-5$ & 0.4976 & 0.4359 & 0.5032 & 0.4451 \\
& $0$  & 0.6409 & 0.4877 & 0.5812 & 0.5930 \\
& $5$  & 0.9608 & 0.9641 & 0.9647 & 0.9717 \\

\bottomrule
\end{tabular}}

\end{minipage}
\end{figure*}

\subsection{Posterior-Weighted Input Embedding (PWE)}

The first step, PWE, incorporates channel information into the MLM input at unreliable token positions. For each vocabulary candidate, the receiver evaluates the symbol likelihoods assuming that the candidate was transmitted, and constructs a token-level channel-posterior. From this, PWE makes a soft-token embedding to use as the MLM input in place of the fixed mask-token embedding from the pre-trained lookup table. This allows the MLM to infer masked tokens using surrounding context of channel-supported candidates.

For an unreliable position $i$ with $m^{(i)}=1$, suppose that vocabulary candidate $v$ was transmitted instead. Let $\mathcal N^{(i)}\subseteq\{1,\ldots,N\}$ denote the indices of transmitted symbols in $\bx$ that carry the bits of token $t^{(i)}$, where the $n$-th symbol may also contain bits from another token. Thus, its likelihood is summed over all possible symbols, $\mathcal X^{(i)}_n(v)\subseteq\mathcal X$. Finally, the likelihood of each candidate token is obtained by combining the likelihoods over all symbols in $\mathcal{N}^{(i)}$, expressed as $\lambda^{(i)}(v) \triangleq \prod\nolimits_{n\in\mathcal{N}^{(i)}} \sum\nolimits_{x\in\mathcal{X}^{(i)}_n(v)} \lambda_n(x)$.
\\The vocabulary-level channel-posterior is obtained by normalizing these likelihoods over all vocabulary candidates of $t^{(i)}$:
\begin{equation}
    q^{(i)}(v) = \frac{\lambda^{(i)}(v)}{\sum_{u\in\mathcal{V}}\lambda^{(i)}(u)}.
\label{eq:channel_post}
\end{equation}
Let $\mathbf{e}(v)\in\mathbb{R}^{d}$ denote the embedding vector of $v$ from the lookup table. The channel-aware soft embedding is defined as a weighted average of token embeddings according to $q^{(i)}(\cdot)$:
\begin{equation}
    \mathbf{e}^{(i)}_{\mathrm{soft}} = \sum_{v\in\mathcal{V}} q^{(i)}(v)\mathbf{e}(v).
\end{equation}
The input embedding provided to the MLM is replaced with soft embedding in unreliable positions to incorporate information of candidates, defined as
\begin{equation}
    \mathbf{e}^{(i)}_{\mathrm{PWE}} = \begin{cases}
    \mathbf{e}^{(i)}_{\mathrm{soft}}, & \text{if } m^{(i)}=1,\\
    \mathbf{e}(\hat t^{(i)}), & \text{if } m^{(i)}=0. \end{cases}
    \label{eq:pwe_input}
\end{equation}
The resulting input embedding sequence of tokens is then fed into the pre-trained MLM, denoted as $F_{\mathrm{MLM}}(\cdot)$, to yield output logits over the vocabulary, $\mathbf{z}^{(i)}_{\mathrm{PWE}}\in\mathbb{R}^{|\mathcal{V}|}$, as
\begin{equation}     
    [\bz^{(1)}_{\mathrm{PWE}}, \ldots, \bz^{(L)}_{\mathrm{PWE}}] = F_{\mathrm{MLM}}([\mathbf{e}^{(1)}_{\mathrm{PWE}},\ldots,\mathbf{e}^{(L)}_{\mathrm{PWE}}]). 
\label{eq:ie_mlm} 
\end{equation}
This is the inference result of jointly considering the channel-aware candidates across the surrounding masked positions.

\subsection{Adaptive Output Logit Fusion (AOF)}
\label{subsec:output_fusion}
After MLM inference, the second step, AOF, further incorporates the channel-posterior into the resulting MLM output before the final token decision to suppress contextually plausible but channel-inconsistent candidates. For the $i$-th token, $\bz^{(i)}_{\mathrm{PWE}}$ is the logit vector from Eq.~\eqref{eq:ie_mlm}, where $z^{(i)}_{\mathrm{PWE}}(v)$ is the logit value assigned to candidate $v\in\mathcal{V}$. To make contextually plausible candidates with low channel posterior less likely in the final inference, the MLM logits are fused as:
\begin{equation}
    z^{(i)}_{\mathrm{AOF}}(v) = z^{(i)}_{\mathrm{PWE}}(v) + \omega\log q^{(i)}(v), \qquad v\in\mathcal{V},
    \label{eq:output_fused_logit}
\end{equation}
where the channel weight $\omega$ controls the contribution of the channel-posterior according to the channel quality, characterized by the estimated bit error rate (BER), $P_{\mathrm{b}}(\hat{\gamma}_{\mathrm{b}})$. This uses estimated SNR, $\hat{\gamma}_{\mathrm{b}}=\frac{E_{\mathrm{s}}}{\hat{N}_0\log_2 M}$, and depends on the given modulation; for example, for BPSK, $P_{\mathrm{b}}(\hat{\gamma}_{\mathrm{b}})=Q(\sqrt{2\hat{\gamma}_{\mathrm{b}}})$, as
\begin{equation}
    \omega \triangleq
    \exp \left[-\left(\frac{P_{\mathrm{b}}(\hat{\gamma_{\mathrm{b}}})}{\tau_{\mathrm{g}}}
    \right)^{\kappa}\right] \in [0,1].
\label{eq:ber_dependent_weight}
\end{equation}
This exponential function reduces $\omega$ from 1 toward 0 as BER grows, where $\tau_{\mathrm{g}}>0$ sets the BER level at which the channel contribution decreases, while $\kappa>0$ controls its rate. $\kappa$ and $\tau_{\mathrm{g}}$ are determined by fitting the function across different SNRs.

The resulting logit vector $\mathbf{z}^{(i)}_{\mathrm{AOF}}$ consists of the candidate-wise value $z^{(i)}_{\mathrm{AOF}}(v)$. The final vocabulary distribution is obtained as $\mathbf{p}^{(i)}_{\mathrm{AOF}}= \mathrm{softmax}\left(\mathbf{z}^{(i)}_{\mathrm{AOF}}\right)$, where $p^{(i)}_{\mathrm{AOF}}(v)$ is the probability assigned to $v$. Accordingly, $p^{(i)}_{\mathrm{AOF}}(v)$ is the final fused probability, assigning higher probability to candidates likely from both MLM inference and the channel-posterior. Finally, the reconstructed token at the $i$-th position becomes
\begin{equation}
    \tilde{t}^{(i)}= \begin{cases}
    \displaystyle
    \arg\max_{v\in\mathcal{V}} p^{(i)}_{\mathrm{AOF}}(v),& \text{if }   m^{(i)}=1,
    \\\hat{t}^{(i)}, & \text{if }   m^{(i)}=0. 
    \end{cases}
\label{eq:final_token_reconstruction}
\end{equation}
The resulting reconstructed token sequence is denoted by
\begin{equation}
    \tilde{\bt}= [\tilde{t}^{(1)}, \tilde{t}^{(2)}, \ldots, \tilde{t}^{(L)}].
\label{eq:reconstructed_token_sequence}
\end{equation}
Thus, unreliable tokens are finally reconstructed from the fused distribution. AOF adjusts each candidate probability according to the channel-posterior, so that candidates with both high MLM prediction probability and high channel-posterior are more likely to be selected.

\begin{figure*}[!t]
\centering
\begin{minipage}[t]{0.33\textwidth} 
\centering
\vspace{0pt}
\includegraphics[width=\linewidth]{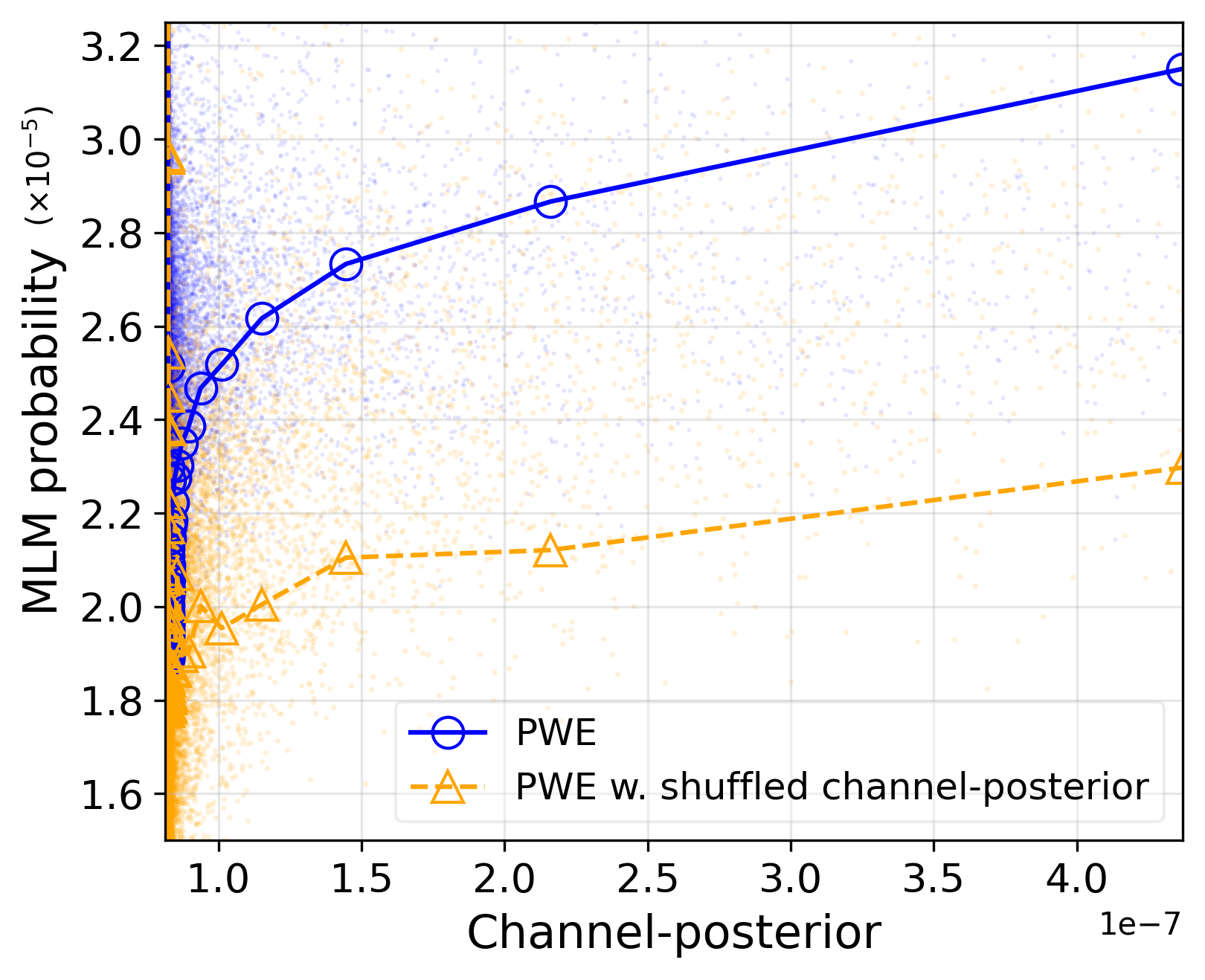}
\captionsetup{width=0.9\linewidth, singlelinecheck=false}
\captionof{figure}{MLM output probability after PWE using channel-posterior and randomly permuted channel-posterior.}
\label{fig:PWE}
\end{minipage}
\hspace{-0.015\textwidth}
\begin{minipage}[t]{0.307\textwidth} 
\centering
\vspace{0pt}
\includegraphics[width=\linewidth]{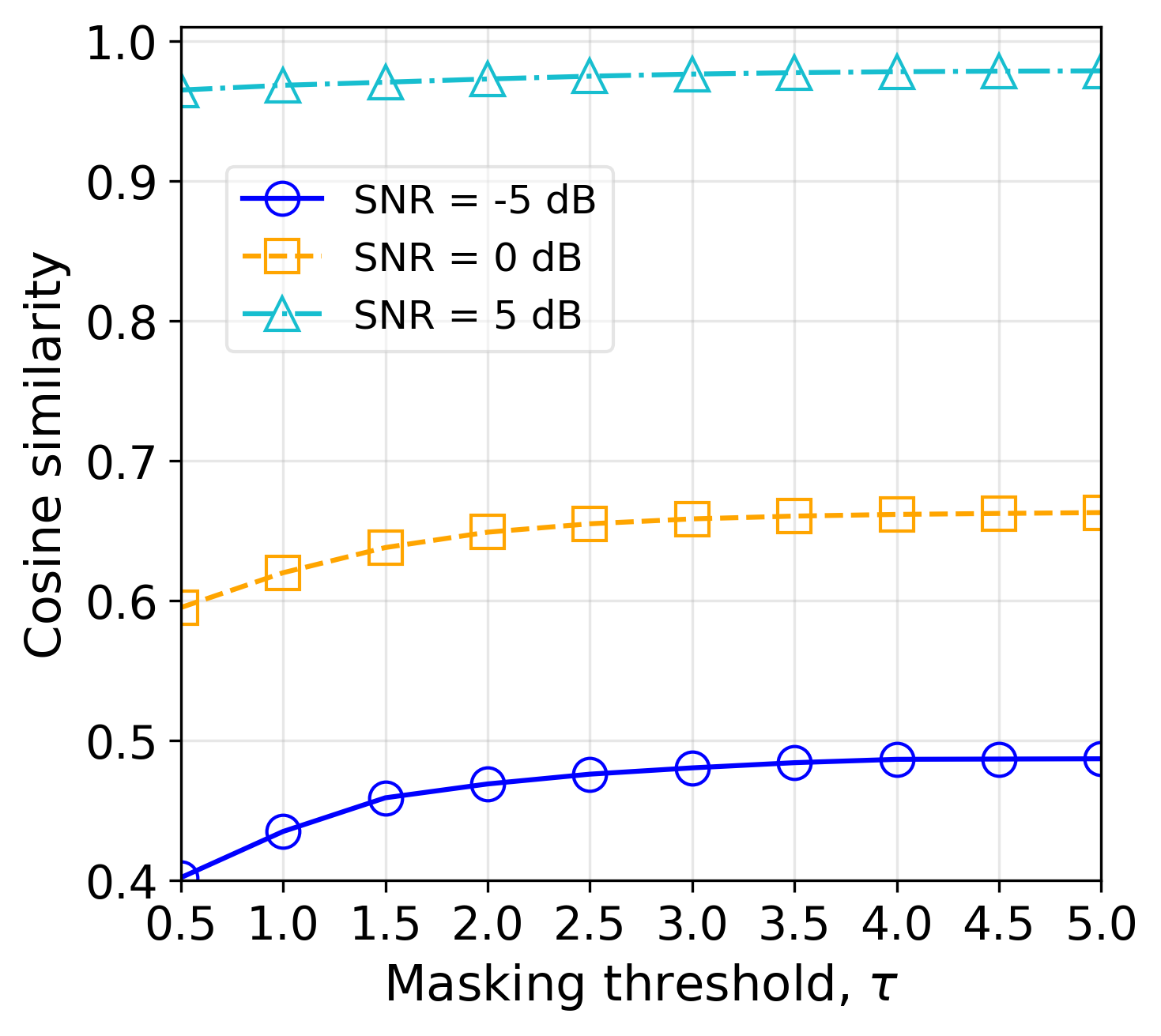}
\captionsetup{width=0.95\linewidth, singlelinecheck=false}
\captionof{figure}{PWE+AOF performance versus the masking threshold; higher values make more tokens unreliable.}
\label{fig:tau}
\end{minipage}%
\hspace{-0.015\textwidth}
\begin{minipage}[t]{0.371\textwidth} 
\centering
\vspace{0pt}
\includegraphics[
    width=\linewidth
]{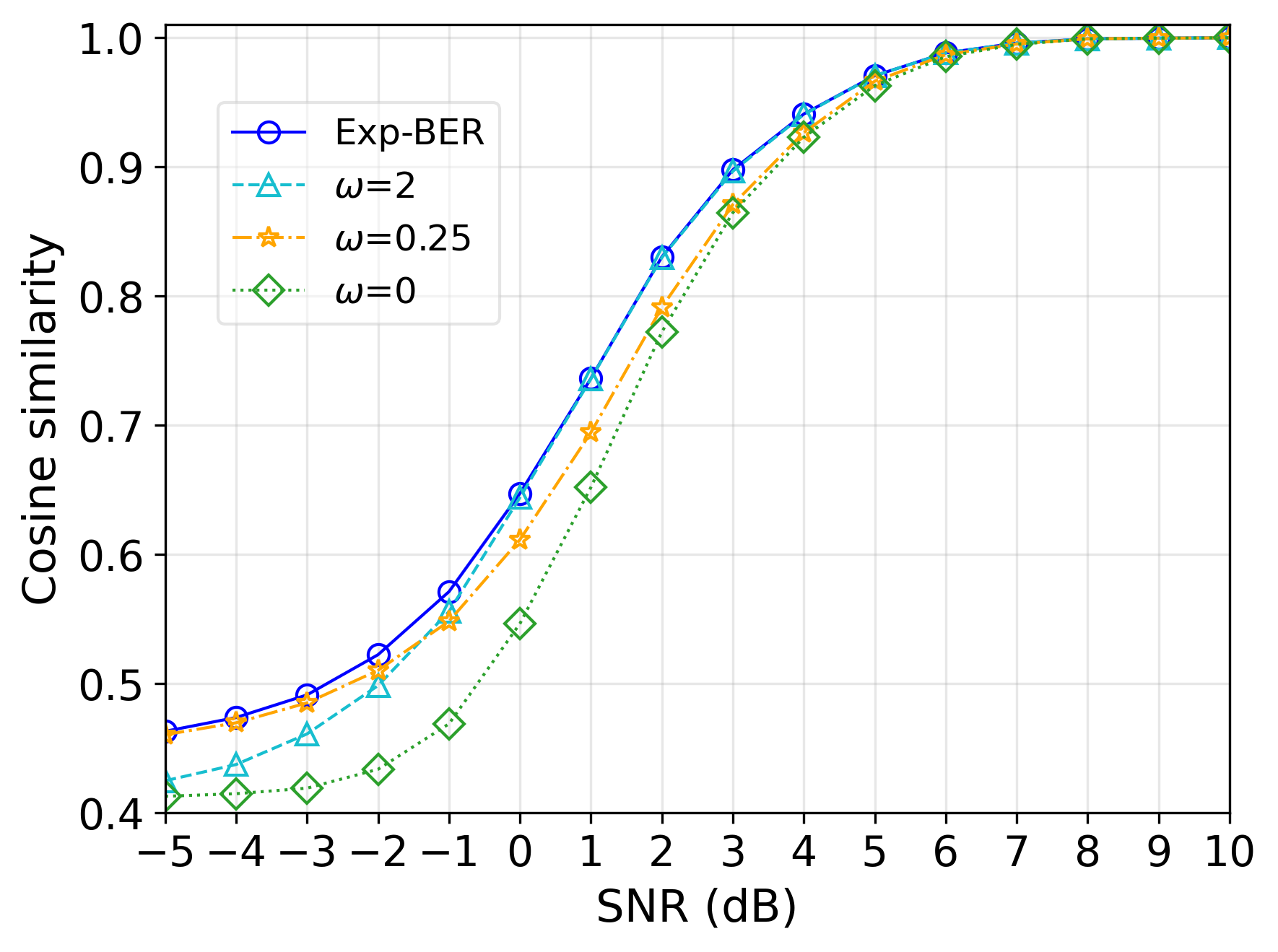}
\captionsetup{width=0.9\linewidth, singlelinecheck=false}
\captionof{figure}{PWE+AOF performance for different fusion weights, $\omega$, including the adaptive values, namely Exp-BER.}
\label{fig:weight}
\end{minipage}
\end{figure*}

\section{Numerical Results}
In this section, we evaluate the proposed two-stage PWE+AOF framework, including the individual contributions of PWE and AOF, against a context-only MLM (C-MLM). For PWE, unreliable tokens are reconstructed from
$\arg\max_{v\in\mathcal{V}} p^{(i)}_{\mathrm{PWE}}(v)$, where $\mathbf{p}^{(i)}_{\mathrm{PWE}}= \mathrm{softmax}\left(\mathbf{z}^{(i)}_{\mathrm{PWE}}\right)$. For AOF,
predefined mask-token embeddings are used, and the channel posterior is fused with the C-MLM output logits.

\subsection{Simulation Settings}
We use $5{,}000$ sentences from the COCO caption dataset~\cite{lin2014microsoft}, tokenized using the BERT tokenizer~\cite{devlin2019bert} with vocabulary size $|\mathcal{V}|=30{,}522$ and $B=15$ bits per token. For MLM, a pretrained BERT-Small~\cite{turc2019wellreadstudentslearnbetter} model with 4 layers, hidden size of 512, and 8 attention heads is used. Transmission is evaluated with SNR $\gamma_{\mathrm{b}}$ from $-5$ to $10$ dB, averaged over 50 independent noise samples using $\mathrm{n}_{\mathrm{p}}=32$ pilots. Reconstruction performance is measured by cosine similarity between sentence embeddings from a pretrained BERT-base encoder~\cite{devlin2019bert}. The masking threshold is set to $\tau=4$. The parameters for $\omega$ calculation are obtained by curve fitting Eq.~\eqref{eq:ber_dependent_weight}, yielding $\tau_{\mathrm{g}}=0.173$ and $\kappa=1.547$.

\subsection{Effect in Different Channels}
Fig.~\ref{fig:modul} shows that the proposed full framework achieves the best reconstruction performance by combining the complementary effects of PWE and AOF. Under the same SNR, 16-QAM has a higher BER than BPSK, yielding less reliable LLRs and distributing the channel-posterior over more token candidates. Consistently in both modulation schemes, PWE yields high reconstruction performance at high SNR as the channel-posterior is concentrated on fewer candidates and therefore provides more discriminative guide to the MLM inference. In contrast, AOF re-ranks the MLM result when channel-posterior is discriminative thus occuring no gain on reconstruction, while at low SNR, less sharp channel-posterior exaggerate the high MLM inference probability with likely candidates. By combining these complementary advantages, PWE+AOF improves cosine similarity over C-MLM by up to $67.2\%$ for BPSK and $46.9\%$ for 16-QAM, at $0$ and $3$ dB, respectively. 
Table~\ref{tab:model} further shows the same tendency across different MLM architectures at representative low, medium, and high SNRs of $-5$, $0$, and $5$ dB. At $-5$ dB, AOF improves the cosine similarity over C-MLM by an average of $21.0\%$. At $5$ dB, PWE improves it by an average of $7.7\%$. Across all evaluated SNRs and MLM architectures, PWE+AOF achieves an average improvement of $26.5\%$ over C-MLM.

Fig.~\ref{fig:PWE} examines how the channel-posterior affects the MLM output through PWE. For comparison, `PWE with shuffled channel-posterior' randomly permutes the posterior values across vocabulary candidates before constructing the soft-token embedding. The scattered dots show individual channel-posterior and MLM output probability pairs, while the curves represent their averages over 40 groups. With PWE, the MLM output probability increases with the channel-posterior, showing that the soft-token embedding guides MLM inference using channel information. This relation is weakened with the shuffled posterior, indicating that PWE is more effective when the channel-posterior and contextual inference assign high probability to the same vocabulary candidates.

\subsection{Effect of Masking Threshold and Channel Weight}
These experiments in this subsection aim to validate the parameter choices of the proposed framework. Fig.~\ref{fig:tau} shows the impact of the masking threshold $\tau$ on the performance. A small $\tau$ leaves unreliable tokens unmasked, whereas increasing $\tau$ masks and reconstructs more unreliable tokens, thereby improving the performance. The gain is larger at low SNR where more token errors occur and they benefit from reconstruction. The improvement is below $0.1\%$ beyond $\tau=4$ in all SNRs, which we set as further experiments.

Fig.~\ref{fig:weight} evaluates the channel weight $\omega$, where $\omega=0$ corresponds to PWE only, while $0.25$ and $2$ assign lower and higher contributions of the channel-posterior relative to the PWE contextual reconstruction. At low SNR, the channel-posterior is less discriminative across token candidates, so a large $\omega$ gives weight to weak candidate probabilities and degrades reconstruction. At high SNR, a larger weight is beneficial as the posterior becomes more reliable. Exp-BER adapts $\omega$ to this SNR-dependent trade-off and achieves the best performance across the SNRs.

\subsection{Comparison with Related Works}
Table~\ref{tab:method_comparison} compares the proposed method with existing LM-aided text reconstruction methods in \cite{hao2025short, wang2026context, wang2025large}. For experiements using LLM and Bidirectional and Auto-Regressive Transformers (BART), we used Pythia-31M~\cite{biderman2023pythia} and BART-Small~\cite{Di_Liello_bart-small}, respectively, for comparable model sizes.

SEC in \cite{hao2025short} feeds the entire errored received text into a BART for contextual reconstruction. Similar to C-MLM, it does not integrate channel information, resulting in lower performance across the SNRs. ICD in \cite{wang2026context} obtains bit-flip candidates and their reliability from symbol-bit (de)mapping Transformers, and combines the LLM output probability with the reliability in a fixed weight. LLM-SC in \cite{wang2025large} also evaluates candidates by combining LLM output and channel-derived likelihoods. These output combination, similar to AOF, performs relatively well among the compared methods in low SNR, but both ICD and LLM-SC still underperforms ours since channel information is not incorporated into LM inference. Additionally, their sequential reconstruction makes early errors to affect subsequent reconstruction, while our method jointly reconstructs multiple errored tokens at once.

Iterative LM-based reconstruction not only suffers from sequential error propagation at low SNR but also incurs substantial computational cost, as summarized in Table~\ref{tab:method_comparison}. A single MLM inference over an $L$-token sequence has $O(L^2)$ complexity due to Transformer self-attention, whereas autoregressive LLM and BART inference sequentially processes increasing sequence lengths, resulting in $\sum_{i=1}^{L}O(i^2)=O(L^3)$. At $\mathrm{SNR}=-5$ dB, where all $L$ tokens are unreliable, PWE+AOF requires one-shot MLM inference and posterior-weighted embedding over the vocabulary into a 512-dimensional vector. LLM-SC and ICD perform LLM inference over $K\leq L$ beams and candidates, respectively, further increasing the computational complexity with $K$. ICL-SEC repeats MLM inference for $R$ refinement iterations and thus becomes more computationally expensive for $R\geq2$ with the average sequence length of $L=32$ in our experiments, while approximately five iterations are reported for convergence \cite{wang2026icl}.

\begin{table}[t]
\centering
\caption{Comparison with language model-aided reconstruction methods.}
\label{tab:method_comparison}

\renewcommand{\arraystretch}{0.9}
\setlength{\tabcolsep}{4pt}
\small

\begin{tabular}{
@{}
>{\centering\arraybackslash}m{0.15\columnwidth}
c c c c c 
@{}
}
\toprule

\footnotesize{Evaluations}
& \makecell{SNR\\(dB)}
& PWE+AOF
& SEC~\cite{hao2025short}
& ICD~\cite{wang2026context}
& LLM-SC~\cite{wang2025large} \\

\midrule

\multirow{3}{*}{\footnotesize{\makecell{Cosine\\Similarity}}}
& $-5$
& 0.4631
& 0.3230
& 0.4068
& 0.3400 \\

& $0$
& 0.6473
& 0.4600
& 0.5298
& 0.5880 \\

& $5$
& 0.9707
& 0.9497
& 0.9620
& 0.9680 \\

\hline

\makecell{{\footnotesize Computation}\\{\footnotesize Complexity}}
& $-5$
& \makecell[t]{\makecell{$\mathcal{O}(L^2$\\$+512L|\mathcal{V}|)$}}
& \makecell[t]{$\mathcal{O}(L^3)$}
& \makecell[t]{\makecell{$\mathcal{O}(KL^3)$\\$+C_{\rm aux}$}}
& \makecell[t]{$\mathcal{O}(KL^3)$} \\

\bottomrule
\end{tabular}

\end{table}

\section{Concluding Remarks}
This paper proposed a dual-stage channel-posterior integration framework for MLM-aided token reconstruction over conventional digital modulation. In the proposed framework, PWE incorporates the channel-posterior into the MLM input through soft-token embeddings, while AOF reuses the same posterior at the output with channel-adaptive weighting. By combining integration at input and output, this framework jointly reconstructs multiple unreliable tokens in a single MLM inference. Numerical results show gains across various SNRs, modulation schemes, and MLM architectures, outperforming existing LM-aided methods with lower complexity.

\end{document}